\documentclass[conference]{IEEEtran}
\IEEEoverridecommandlockouts

\usepackage{graphicx} 
\usepackage{pifont}
\usepackage{amsmath}

\usepackage{amssymb}  
\usepackage{amsfonts}
\usepackage{algorithmic}
\usepackage{array}
\usepackage{multicol}
\usepackage{mathrsfs}
\usepackage{textcomp}
\usepackage{stfloats}
\usepackage{url}
\usepackage{verbatim}
\usepackage{graphicx}
\usepackage{caption}
\usepackage{subcaption}
\usepackage[ruled, vlined, linesnumbered]{algorithm2e}
\usepackage{cite}
\usepackage{booktabs}
\usepackage{amssymb,amsfonts,bm}
\usepackage{amsmath,tikz}
\usepackage{color}
\usepackage{mathtools}
\usepackage[hidelinks]{hyperref} 
\usepackage{rotating}
\usepackage{blkarray}
\usepackage{physics}
\usetikzlibrary{arrows}
\usepackage{makecell}

\usepackage{amsthm}
\usepackage{comment}
\usepackage{multirow}
\usepackage{geometry}
\graphicspath{{images/}}

\begin{document}

\title{\LARGE \bf
Swept Volume-Aware Trajectory Planning and MPC Tracking for Multi-Axle Swerve-Drive AMRs \\

\author{
Tianxin Hu$^1$, 
Shenghai Yuan$^{1*}$,~\textit{Member,~IEEE},\\ 
Ruofei Bai, 
Xinhang Xu, 
Yuwen Liao,
Fen Liu,
Lihua Xie,~\textit{Fellow,~IEEE}

\thanks{This research is supported by the National Research Foundation, Singapore, under its Medium-Sized Center for Advanced Robotics Technology Innovation (CARTIN). }
\thanks{All authors are with the School of Electrical and Electronic Engineering, Nanyang Technological University, 50 Nanyang Avenue, Singapore 639798, 
   { Email:  \{shyuan, elhxie\}@ntu.edu.sg, tianxin001@e.ntu.edu.sg.}}%
}
\thanks{$^*$ Corresponding Author, $^1$ Equal contribution. }

}

\maketitle

\begin{abstract}
Multi-axle autonomous mobile robots (AMRs) are set to revolutionize the future of robotics in logistics. As the backbone of next-generation solutions, these robots face a critical challenge: managing and minimizing swept volume during turns while maintaining precise control. Traditional systems designed for standard vehicles often struggle with the complex dynamics of multi-axle configurations, leading to inefficiency and increased safety risk in confined spaces.
Our innovative framework overcomes these limitations by combining swept volume minimization with Signed Distance Field (SDF) path planning and model predictive control (MPC) for independent wheel steering. This approach not only plans paths with an awareness of the swept volume, but actively minimizes it in real-time, allowing each axle to follow a precise trajectory while significantly reducing the space the vehicle occupies. By predicting future states and adjusting the turning radius of each wheel, our method enhances both maneuverability and safety, even in the most constrained environments.
Unlike previous works, our solution goes beyond basic path calculation and tracking, offering real-time path optimization with minimal swept volume and efficient individual axle control. To our knowledge, this is the first comprehensive approach to tackle these challenges, delivering life-saving improvements in control, efficiency, and safety for multi-axle AMRs. Furthermore, we will open-source our work to foster collaboration and enable others to advance safer and more efficient autonomous systems.
\end{abstract}


\section{Introduction}
Logistics is a vital component of modern society, which facilitates the flow of goods and services. Although large multi-axle trucks \cite{zhao2019modelling,islam2020planning}  offer capacity and efficiency, they face several significant maneuverability challenges \cite{yuan2021survey} in constrained urban environments.

\begin{figure}
    \centering
    \includegraphics[width=1\linewidth]{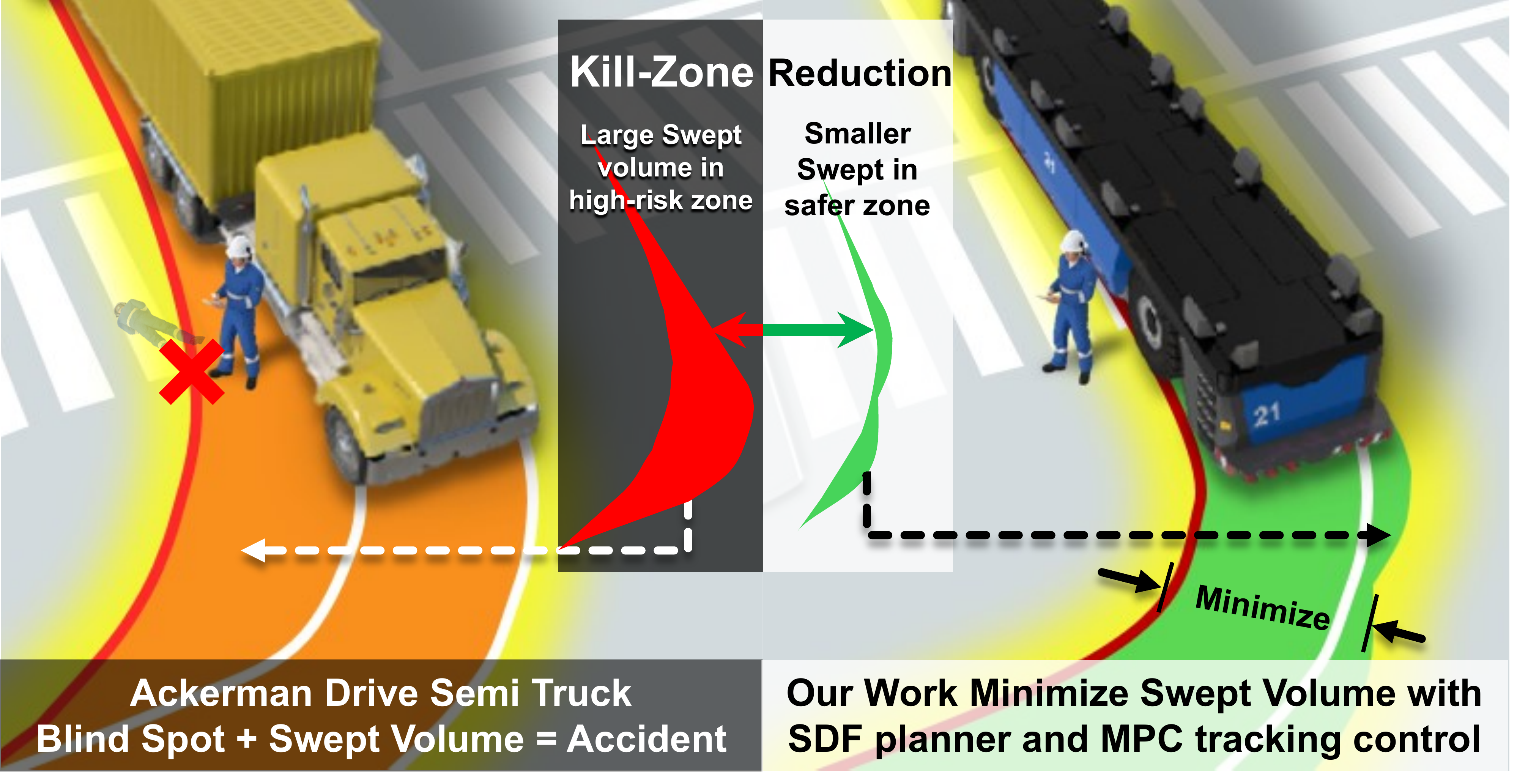}
    \caption{\footnotesize This work aims to reduce the minimal swept volume and ensure stable trajectory tracking, enhancing safety in industrial applications..}
    \label{fig:firstpage}
    \vspace{-15pt}
\end{figure}

A critical challenges is the swept volume \cite{zhang2023continuous,wang2024implicit} of multi-axle vehicles, as the rear axles follow different paths from the front during turns \cite{liu2020experimental}, as shown in Fig. \ref{fig:firstpage}. This increases the risk of entering adjacent lanes, sidewalks, or obstacles. Current automated driving systems are insufficient for addressing this problem, as they are primarily designed for smaller vehicles \cite{yang2024e2e} (i.e., cars) and fail to optimize for the unique dynamics of multi-axle configurations \cite{zhao2019modelling}.

The challenge lies in managing the swept volume while ensuring that each axle follows a safe and efficient path in real-time \cite{wang2024implicit,zhang2023continuous}. Existing methods for path planning \cite{bai2021multi,bian2023risk,jin2024gs,liu2023safe,cao2023path,cao2023neptune,li2024jacquard,zhao2024design,yang2024trace,fan2024flying,yu2024aggressive,Bai2025Realm} and control \cite{cao2019bearing,liu2023non,hu2023stackelberg,cao2020online,liu2023multiple,xu2024cost,er2013development,jia2023evolver,liu2024distance,ji2024integration} typically simplify the vehicle as a single rigid body \cite{lee2022cyclops}, which does not account for the independent control needed for each axle in multi-axle systems. As a result, these approaches \cite{bai2021multi,bian2023risk,jin2024gs,liu2023safe,cao2023path,cao2023neptune,li2024jacquard,zhao2024design,yang2024trace,fan2024flying,yu2024aggressive,Bai2025Realm,cao2019bearing,liu2023non,hu2023stackelberg,cao2020online,liu2023multiple,xu2024cost,er2013development,jia2023evolver,liu2024distance,ji2024integration} struggle to minimize swept volume, especially in complex environments.

To address this challenge, we propose a novel approach that integrates swept volume-aware path planning with model predictive control (MPC) for Swerve-Drive autonomous mobile robots (AMR) systems. Our method optimizes the trajectory of the vehicle in real-time, minimizing the swept volume while allowing each wheel to turn independently. By utilizing MPC, we can predict the future state of the vehicle and adjust the turning radius of each wheel, ensuring precise navigation through tight spaces without compromising safety or efficiency.

This approach not only tackles the limitations of current systems, but also sets the foundation for safer and more robust multi-axle AMR operations in real-world environments. By reducing swept volume and improving control over each axle, our method paves the way for more reliable autonomous heavy vehicles, particularly in logistics and industrial applications.

The main contributions of this work are as follows.

\begin{itemize}
    \item We propose a unified approach that integrates swept volume-aware path planning with MPC to optimize the trajectory and independently control each wheel of multi-axle AMRs, ensuring precise maneuverability in constrained environments.

    \item We propose a method for calculating the steering angles of each wheel group in multi-axle vehicles based on velocity vectors, simplifying the vehicle model and facilitating the use of MPC control.

    \item We validate the approach, showing significant reductions in swept volume and improved real-time trajectory tracking using CUDA, enabling more reliable and efficient autonomous heavy-duty AMR applications.

    \item We will open-source our work for the benefit of the community. \url{https://github.com/ccwss-maker/svplan}
\end{itemize}

%%%%%%%%%%%%%%%%%%%%%%%%%%%%%%%%%
\begin{figure*}[t]
    \centering
    \vspace{-15pt}
    \includegraphics[width=0.8\linewidth]{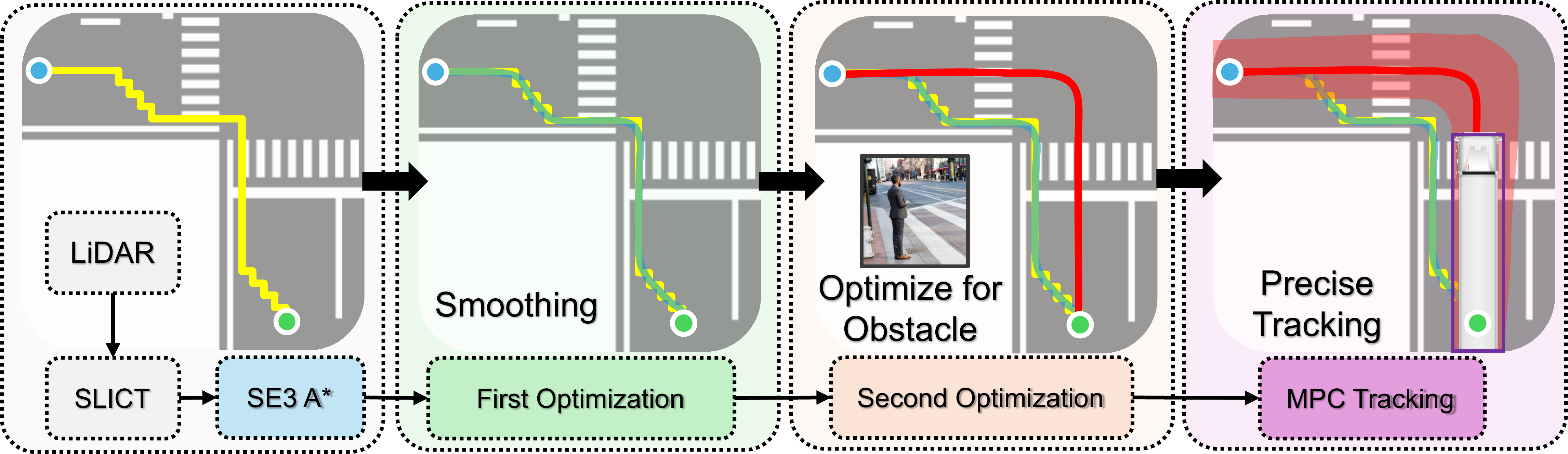}
    \caption{ \footnotesize The proposed solution uses LiDAR inertial odometry \cite{nguyen2024eigen} for front-end odometry, with multi-stage back-end planning and MPC to minimize swept volume iteratively.
 }
    \label{fig:process}
    \vspace{-15pt}
\end{figure*}
%%%%%%%%%%%%%%%%%%%%%%%%%%%%%%%%%%
\vspace{-6pt}
\section{Related Work}
\vspace{-5pt}
In trajectory planning, swept volume detection is critical for obstacle avoidance of semi-trucks and service AMRs, yet few works address this issue comprehensively \cite{zhao2019modelling,liu2020experimental,baxter2020deep,chiang2021fast,zips2015optimisation, cao2019preview}. Basic methods, such as \cite{ilic2018vehicle}, rely on GPS for odometry \cite{chen2024salient,esfahani2019deepdsair,yang2024fast,10612831,ji2022robust,Li2024graph,esfahani2018new,10801455,esfahani2020local,10802614,jin2024robust,10802691,li2024hcto,Nguyen2025ULOC,li2024ua,yin2024outram,ji2024sgba,esfahani2021learning,esfahani2019towards,lyu2023spins} and body frame integration, leading to significant calculation errors. Täubig et al. \cite{Täubigsweptvolume} propose a two-stage approach, combining broad-phase detection for quick collision identification \cite{yuan2014Autonomous} with narrow-phase detection \cite{wang2015automatic,wang2017heterogeneous} using the GJK algorithm for precise calculations \cite{wu2019depth}. Although this method balances speed and accuracy, it performs poorly in complex environments compared to the SDF-based approach \cite{WangSDF}. However, the SDF-based approach \cite{WangSDF,wang2024implicit,zhang2023continuous} often lacks real-time performance, suffers from heading tracking issues, and does not integrate traffic rules, which limits its applicability in more practical scenarios.

For trajectory tracking \cite{lyu2021vision,qi2024air}, controlling the steering angles of each wheel in multi-axle Swerve-Drive AMRs in real-time is a highly complex task. While dynamic models with varying degrees of freedom have been studied \cite{xu2022hierarchical, wang2023multi, gao2014turning, zhang2022dual, hu2016control, wu2017optimizing}, these works often overlook the geometric aspects of steering, failing to fully utilize the flexibility of multi-axle independent steering. To address this, models such as Ackerman Steering \cite{chaudhuri2009kinematic,wu2021learn}, Active Front and Rear Steering (AFRS) \cite{ye2016steering}, and Front Wheel Steering (FWS) \cite{xu2022hierarchical} have been proposed, but their steering centers are typically constrained along the first, last or middle axles, limiting maneuverability. In contrast, D-based steering models \cite{zhang2016study, zhang2015steering, zhang2015steering1} allow the steering center to be distributed more flexibly, enhancing steering performance. However, transitioning between different steering modes depending on road conditions still introduces transient disturbances.

\section{Proposed Solution}

\subsection{Problem Statement}
Let $\xi \in \mathbb{R}$ represent the two-dimensional top-view area of the AMR, assuming the vehicle shape remains constant during motion. The objective of this work is to minimize the swept volume $\mathbb{V} = \cup_{t \in [t_{s}, t_{e}]} \xi \mathcal{L}(t) h$, where $h \in \mathbb{R}$ is the constant height of the vehicle and $\mathcal{L}(t)$ denotes the trajectory. The swept volume is defined as the three-dimensional volume swept by the vehicle during its motion. For computational efficiency, we neglect the vehicle's lateral tilt. Since $h$ is constant, minimizing $\mathbb{V}$ is equivalent to minimizing the swept area $\mathbb{S}$, which is the two-dimensional top-view projection of $\mathbb{V}$. This minimization is achieved by optimizing the trajectory control points $\mathcal{L} = \left\{ \left( x_j, y_j, \varphi_j \right) \in \mathbb{SE}(2) \mid j = 1, 2, \dots, N-1 \right\}$ and implementing an MPC controller with control input $u = [V_x, V_y, \omega]^\top \in \mathbb{R}^3$, where $V_x$, $V_y$, and $\omega$ represent the longitudinal, lateral, and rotational velocities of the AMR, respectively.

% The shape $\xi(t)$ is assumed to be static, changing only when the AMR is loaded with a container. For simplicity, we denote it as $\xi$.

\subsection{Trajectory Planning}
As shown in Fig.~\ref{fig:process}, the trajectory planning process consists of four steps. First, the A* algorithm is used to generate an initial feasible path. Then, based on this path, a corresponding heading sequence is estimated. Together, the path and heading sequence form the initial trajectory, denoted by $\mathcal{L}^{A*}(t)$. Subsequently, $\mathcal{L}^{A*}(t)$ is utilized in the first optimization step to generate an initial smoothed continuous trajectory $\mathcal{L}^{F}_{M}(t)$. Finally, $\mathcal{L}^{F}_{M}(t)$ is further optimized in the second optimization step to avoid obstacles and minimize the swept volume $\mathbb{S}$, resulting in the final continuous trajectory $\mathcal{L}^{S}_{M}(t)$.

The superscript in $\mathcal{L}^{F}_{M}(t)$ and $\mathcal{L}^{S}_{M}(t)$ denotes the first and second optimization steps, respectively, where both trajectories are represented using the minimum control effort polynomial trajectory class (MINCO) \cite{wang2022geometrically}. The MINCO is designed to optimize a trajectory by minimizing the control effort while fitting a set of $N$ discrete path points into $N-1$ continuous polynomial segments. Each polynomial segment $P_j(t)$ represents a quintic curve and is optimized over time durations $T_j$. The complete trajectory is represented by:
\[
\mathcal{L}_M = \left\{ P_j(t) : [0, T_j] \to \mathbb{R}^3 \mid j = 1, 2, \dots, N-1 \right\},
\]
where each segment $P_j(t)$ is described by a quintic polynomial:
\[
P_j(t) = C_{0,j} + C_{1,j} t + C_{2,j} t^2 + C_{3,j} t^3 + C_{4,j} t^4 + C_{5,j} t^5.
\]
The coefficients $C_{i,j} \in \mathbb{R}^{3}$ are determined by the control points of the trajectory $q = [q_1, q_2, \dots, q_{N-1}] \in \mathbb{R}^{(N-1)\times3}$ and the for each polynomial segment $T = [T_1, T_2, \dots, T_{N-1}]\in \mathbb{R}^{N-1}$, as defined by the mapping function:
\[
C_{i,j} = M(i, j, q, T),
\]
where:
$i~\in~[0, 5]$ denotes the index of coefficients in $C_{i,j}$ for the j-th component.

\subsubsection{First Optimization}
The first optimization is intended for smoothing of the trajectory by minimizing energy consumption and time consumption of the trajectory while ensuring that the trajectory follows $\mathcal{L}^{A*}(t)$. The optimization problem is defined as:
\begin{equation}
\min_{q, T} \; \mathbb{W}_E \cdot J_E + \mathbb{W}_T \cdot J_T + \mathbb{W}_P \cdot J_P,
\end{equation}
% \[
% \min_{q, T} J_F(c, T), \quad \text{s.t.} \quad C_{i,j} = M(q_i, T_j), 
% \]
% % where $ j$ = 1, 2, $\dots$, N-1 , $q = [q_1, q_2, \dots, q_{N-1}]$ are the control points, $T = [T_1, T_2, \dots, T_{N-1}]$ are the time duration for each segment. 
% where the objective function is defined as follows:
% \begin{equation}
% J_F = \mathbb{W}_E \cdot J_E + \mathbb{W}_T \cdot J_T + \mathbb{W}_P \cdot J_P,
% \end{equation}
where \( \mathbb{W}_E \), \( \mathbb{W}_T \), and \( \mathbb{W}_P \) are the weights for energy consumption, time consumption, and trajectory deviation, respectively, and \( J_E \), \( J_T \), and \( J_P \) represent energy loss, time loss, and trajectory deviation loss. 
%\(J_E\) and \(J_T\), as well as their gradients with respect to \( C_{i,j} \) and \( T_j \), i.e., \( \partial J_E / \partial T_j \), \( \partial J_E / \partial C_{i,j} \), \( \partial J_T / \partial T_j \), and \( \partial J_T / \partial C_{i,j} \), have been calculated in this paper \cite{wang2024implicit}, and will not be discussed further here. 
The gradients of \(J_E\) and \(J_T\) with respect to \(C_{i,j}\) and \(T_j\), including \( \partial J_E / \partial T_j \), \( \partial J_E / \partial C_{i,j} \), \( \partial J_T / \partial T_j \), and \( \partial J_T / \partial C_{i,j} \), have been rigorously derived in \cite{wang2024implicit} and will not be revisited in this paper. \(J_P\) and its gradients are shown as follows:
\begin{equation}
J_P = \sum_{j=1}^{N-1} (P_j - P_{A^*, j})^2,
\end{equation}
\begin{equation}
\frac{\partial J_P}{\partial P_j} = 2(P_j - P_{A^*, j}),
\end{equation}
\begin{equation}
\frac{\partial J_P}{\partial T_j} = \frac{\partial J_P}{\partial P_j} \cdot \frac{\partial P_j}{\partial T_j} = \frac{\partial J_P}{\partial P_j} \cdot V_j.
\end{equation}
Here, \( P_{A^*, j} \) represents the initial trajectory points in $\mathcal{L}^{A*}(t)$, and \( V_j = [V_X, V_Y, \omega]^T \) is the velocity matrix at the corresponding trajectory point, where \(V_X\) and \(V_Y\) are the velocities in the vehicle  \(X\)- and \(Y\)-coordinates, and \( \omega \) is the angular velocity. The optimization is performed using the LBFGS \cite{coppola2020lbfgs}, resulting in the initial optimized trajectory $\mathcal{L}^{F}_{M}(t)$.

\subsubsection{Second Optimization}
\begin{figure}
    \centering
    \includegraphics[width=1\linewidth]{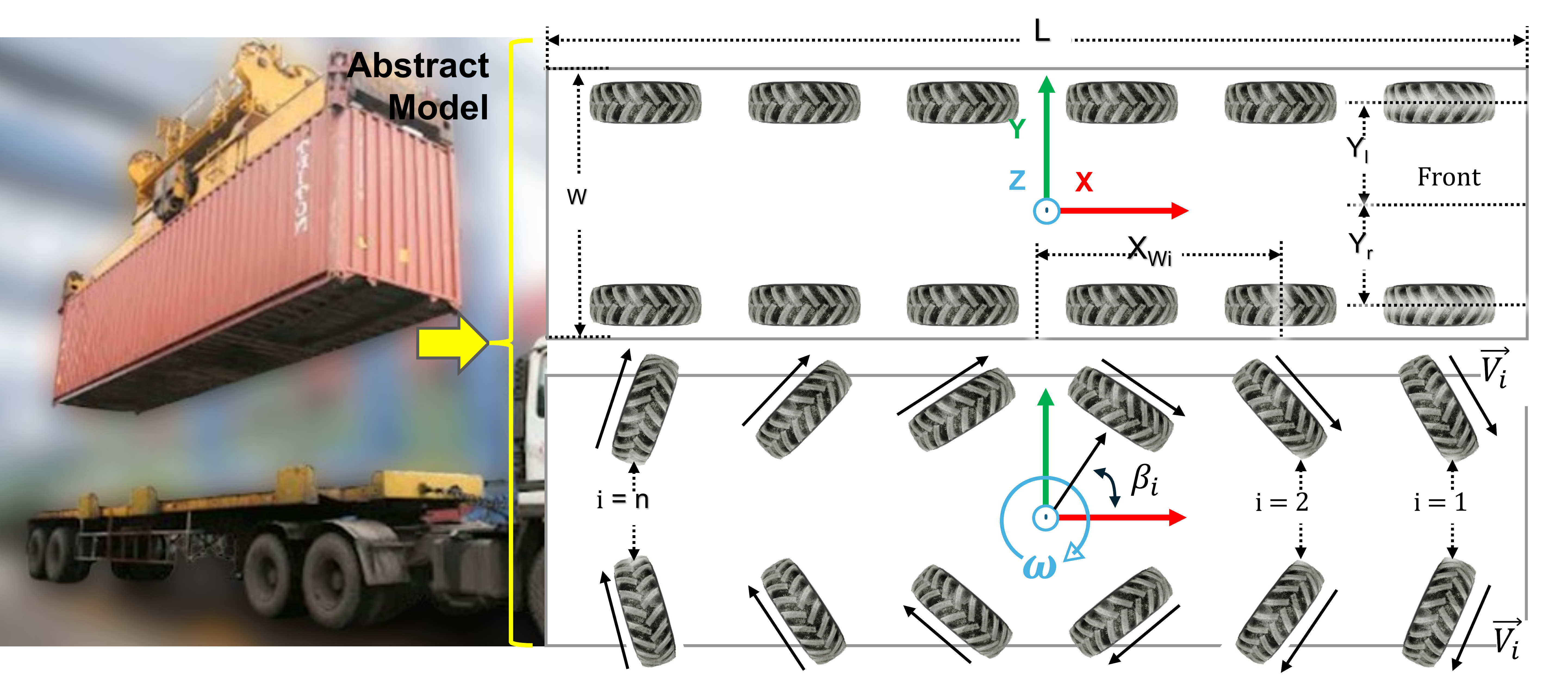}
    \caption{\footnotesize Vehicle Parametric Model}
    \label{fig:modeling}
    \vspace{-15pt}
\end{figure}

In this stage, the optimization objectives are to minimize energy, total time, and safety distance, and to reduce the size of the swept area. Therefore, the optimization problem is defined as:
\begin{equation}
\min_{q, T} \; \mathbb{W}_E \cdot J_E + \mathbb{W}_T \cdot J_T + \mathbb{W}_{ob} \cdot J_{ob} + \mathbb{W}_{sv} \cdot J_{sv}. 
\end{equation}
Here, \( \mathbb{W}_E \), \( \mathbb{W}_T \), \( J_E \), and \( J_T \) are the same as those in the previous stage. \( \mathbb{W}_{ob} \) and \( \mathbb{W}_{sv} \) represent the weights for obstacle safety distance and swept area, respectively, while \( J_{ob} \) and \( J_{sv} \) are their corresponding cost functions. The cost function \( J_{ob} \) and its partial derivatives with respect to \( P_j \) and \( T_j \), i.e., \( \partial J_{ob} / \partial T_j \) and \( \partial J_{ob} / \partial P_j \), are calculated using the Signed Distance Field (SDF) \cite{zhang2023continuous}. In the top-down view as shown in Fig.~\ref{fig:modeling}, the vehicle is approximated as a rectangle. The vehicle coordinate system is defined with its origin located at the geometry center of the vehicle. The X- and Y-axes are aligned with the vehicle's longitudinal and lateral directions, respectively. For a point $P_{veh} = [X_{veh}, Y_{veh}]^T$ in the vehicle coordinate system, its distances to the vehicle's boundary, denoted as $dx$ and $dy$, are given by:
\begin{equation}
dx = \lvert X_{veh} \rvert - \frac{L}{2}, \quad dy = \lvert Y_{veh} \rvert - \frac{W}{2},
\end{equation}
where \( L \) and \( W \) denote the length and width of the vehicle, respectively. Therefore, the SDF function \( f_{SDF}(P_{veh}) \) is defined as:
\begin{equation}
\begin{cases} 
\sqrt{dx^2 + dy^2}, & \text{if } dx > 0 \text{ and } dy > 0, \\
\max(dx, dy), & \text{otherwise}.
\end{cases}
\label{eq:fsdf}
\end{equation}
Thus, the gradient of the SDF, \( \nabla f_{\text{SDF}}(P_{\text{veh}}) \), is computed as follows:
\begin{equation}
\begin{cases}
\left( \frac{dx\cdot \text{sign}(X_{\text{veh}})}{\sqrt{dx^2 + dy^2}}, \frac{dy\cdot \text{sign}(Y_{\text{veh}})}{\sqrt{dx^2 + dy^2}} \right), & \text{if } dx > 0 \text{ and } dy > 0, \\
\left( \text{sign}(X_{\text{veh}}), 0 \right), & \text{else if } dx \geq dy, \\
\left( 0, \text{sign}(Y_{\text{veh}}) \right), & \text{otherwise}.
\end{cases}
\end{equation}
Therefore, when the vehicle is at the $j$-th control point, the obstacle $P_{\text{ob}} = [X_{\text{ob}}, Y_{\text{ob}}]^T$ in the global coordinate system (with its origin at the first control point) is characterized by the vehicle's SDF and gradient as:
\begin{equation}
F_{\text{SDF}}(P_{ob}, j) = f_{\text{SDF}}(R^{-1}(j)(P_{\text{ob}} - T(j)),
\end{equation}
\begin{equation}
\nabla F_{\text{SDF}}(P_{ob}, j) = R(j) \nabla f_{\text{SDF}}(R^{-1}(j)(P_{\text{ob}} - T(j))),
\end{equation}
where $R(j)$ and $T(j)$ represent the rotation matrix and translation vector of the AMR from the first to the $j$-th control point, respectively. The cost function \( J_{\text{ob}} \) is defined as follows:
\begin{equation}
J_{\text{ob}} = \sum_{j=1}^{N-1} \sum_{k=1}^{N_{ob}} J^{'}_{\text{ob}}(k, j),
\end{equation}
where \( k \) represents the obstacle index, and \(N_{ob}\) represents the number of obstacles.  The cost function \( J^{'}_{\text{ob}}(k, j) \) is defined as follows:
\begin{equation}
\begin{cases}
(d_{\text{th}}-F_{\text{SDF}}(P_{\text{ob}}(k), j))^3, & \text{if } F_{\text{SDF}}(P_{\text{ob}}(k), j) < d_{\text{th}}, \\
0, & \text{otherwise},
\end{cases}
\end{equation}
where \( d_{\text{th}} \) represents the safety distance threshold.
In conclusion, the gradients of \( J_{\text{ob}} \) with respect to \( P_j \) and \( T_j \), i.e., \( \partial J_{\text{ob}} / \partial P_j \) and \( \partial J_{\text{ob}} / \partial T_j \), are given by:
\begin{equation}
\frac{\partial J_{\text{ob}}}{\partial P_j} = \sum_{k=1}^{N_{ob}}\nabla F_{\text{SDF}}(P_{\text{ob}}(k, j)),
\end{equation}
\begin{equation}
\frac{\partial J_{\text{ob}}}{\partial T_j} = \frac{\partial J_{\text{ob}}}{\partial P_j} \cdot \frac{\partial P_j}{\partial T_j} = \frac{\partial J_{\text{ob}}}{\partial P_j} \cdot V_j.
\end{equation}
For the swept area, ensuring that the shortest side of the vehicle is always perpendicular to the trajectory, i.e., the long axis of the vehicle is tangent to the trajectory, can reduce the swept area. Therefore, the cost function \( J_{\text{sv}} \) and its gradients are given as follows:
\begin{equation}
\Delta \varphi_j =  \varphi_j - \arctan\left(\frac{V_{Y, j}}{V_{X, j}}\right),
\end{equation}
\begin{equation}
J_{\text{sv}} =  \sum_{j=0}^{N_{\text{p}}} (\Delta \varphi_j)^2,
\end{equation}
\begin{equation}
\frac{\partial J_{\text{sv}}}{\partial \varphi_j} = 2 \Delta \varphi_j,
\end{equation}
\begin{equation}
\frac{\partial J_{\text{sv}}}{\partial X_j} = 2 \Delta \varphi_j \left( \frac{- V_{Y, j}}{V_{X, j}^2 + V_{Y, j}^2} \right),
\end{equation}
\begin{equation}
\frac{\partial J_{\text{sv}}}{\partial Y_j} = 2 \Delta \varphi_j \left( \frac{V_{X, j}}{V_{X, j}^2 + V_{Y, j}^2} \right),
\end{equation}
\begin{equation}
\frac{\partial J_{\text{sv}}}{\partial T_j} = \frac{\partial J_{\text{sv}}}{\partial \varphi_j} \cdot \omega_j + \frac{\partial J_{\text{sv}}}{\partial X_j} \cdot V_{X, j} + \frac{\partial J_{\text{sv}}}{\partial Y_j} \cdot V_{Y, j},
\end{equation}
where \( \omega_j \) represents the yaw rate, and \( V_{X, j} \) and \( V_{Y, j} \) represent the velocities of AMR in the X- and Y-directions at the \( j \)-th control point, respectively. The final trajectory $\mathcal{L}^{S}_{M}(t)$ is obtained by optimizing using the LMBM \cite{karmitsa2020limited}.

\subsubsection{Swept Area Estimation}
From Equation (\ref{eq:fsdf}), the SDF of a point \( P \) in space with respect to the vehicle's trajectory at time \( t \) is given by:
\begin{equation}
F_{\text{SDF}}(P, t) = f_{\text{SDF}}(R^{-1}(t)(P - T(t))),
\label{eq:fsdft}
\end{equation}
where \( R(t) \) and \( T(t) \) represent the vehicle’s rotation matrix and translation matrix from time 0 to time \( t \). To compute the swept-volume SDF of the vehicle along its trajectory, for each point on the map, we compute its minimum distance to the vehicle's swept area by evaluating all vehicle poses from initial time $t_{\text{min}}$ to final time $t_{\text{max}}$. Since for any point, there exists a unique time instant $t^*$ at which the minimum distance occurs, the Armijo line search method \cite{armijo1966minimization} can be employed to find this optimal time $t^*$. By substituting \( t^* \) into Equation (\ref{eq:fsdft}), we can obtain the minimum distance \( f_{\text{SDF}}^* = F_{\text{SDF}}(P, t^*)\) at a point $P$.

To calculate the SDF for the swept area, an \( \mathbb{N} \times \mathbb{N} \) grid map is constructed over the region of interest. For each of the \( \mathbb{N}^2 \) grid points, multiple iterations of the Armijo line search method are required to determine the optimal time $t^*$. Once \( t^* \) is obtained, it is then used to calculate \( f_{\text{SDF}}^* \). This process is computationally intensive and time-consuming. However, using CUDA, we can assign \( \mathbb{N}^2 \) threads to simultaneously compute \( t^* \) for all grid points, greatly accelerating the computation.

%\vspace{-0.25em}

\subsection{Trajectory Tracking}

\begin{figure}
    \centering
    \includegraphics[width=0.9\linewidth]{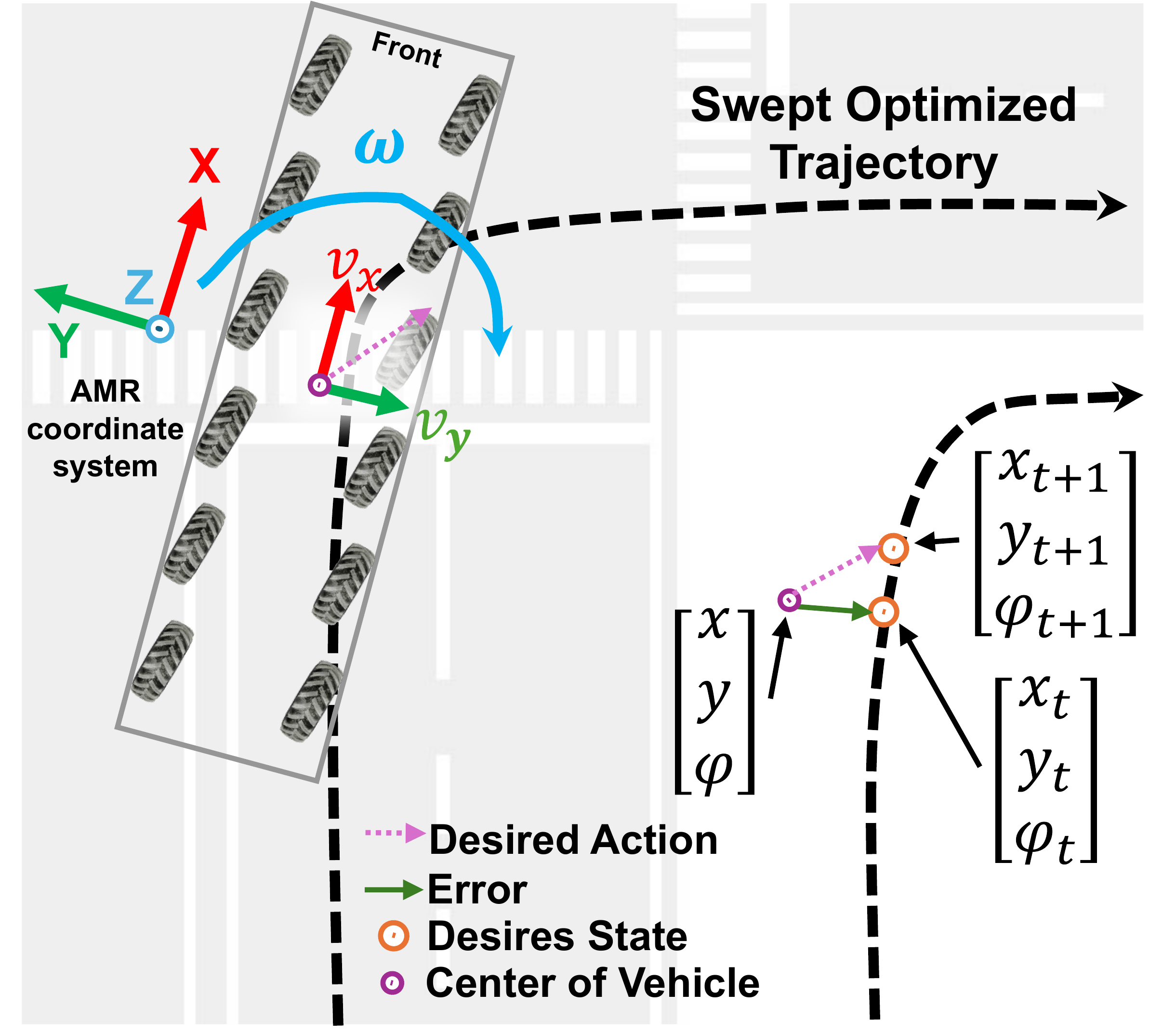}
    \vspace{-5pt}
    \caption{\footnotesize Multi-Axle AMR MPC tracking control.}
    \label{fig:MPCcontrol}
    \vspace{-15pt}
\end{figure}

%As shown in Figure~\ref{fig:1}, a Cartesian coordinate system is established with the geometric center point of the vehicle as the origin. The X-axis and Y-axis represent the longitudinal and lateral axes of the vehicle, respectively. The vehicle length is denoted as \( L \), and the wheelbase is denoted as \( W \). The coordinate of the i-th wheel on the X-axis is \( X_{wi} \in \left(-\frac{L}{2}, \frac{L}{2}\right) \), and on the Y-axis, the coordinates of the left and right wheels are \( Y_{wi} \in \left\{ -\frac{W}{2}, \frac{W}{2} \right\} \).
As shown in Fig.~\ref{fig:MPCcontrol}, the $i$-th wheel is located at $X_{wi} \in \left(-{L}/{2}, {L}/{2}\right)$ along the vehicle's X-axis, with the left and right wheels positioned at $Y_{wi} \in \left\{ -{W}/{2}, {W}/{2} \right\}$ along the Y-axis.
The state vector is chosen as \( \mathbf{X} = [x, y, \varphi]^T \), and the control vector is \( u = [V_{x}, V_{y}, \omega]^T \). The discretized vehicle control state-space equation can be obtained as:
\begin{equation}
 \mathbf{X}(k+1) = A\mathbf{X}(k)+Bu(k),
\label{eq:3}
\end{equation}
where \(A\) is identity matrix, \(B\) is a diagonal matrix with \(T\) on its main diagonal. Therefore, based on the recursive application of Equation (\ref{eq:3}), the state vector of the vehicle from time step \( k+1 \) to \( k+N_p \) can be obtained and is expressed as follows:
\begin{equation}
Y = \Psi \mathbf{X}(k) + \Theta U,
\label{eq:4}
\vspace{-5pt}
\end{equation}
where \( N_p \) represents the prediction horizon, \( N_c \) represents the control horizon, and \( Y \), \( \Psi \), \( U \), and \( \Theta \) are expressed as follows:
\vspace{-10pt}
\[
Y =
\begin{bmatrix}
\mathbf{X}(k+1) \\
\mathbf{X}(k+2) \\
\vdots \\
\mathbf{X}(k+N_p)
\end{bmatrix}
,
\Psi = 
\begin{bmatrix}
A \\
A^2 \\
\vdots \\
A^{N_p}
\end{bmatrix}
, 
 U = 
\begin{bmatrix}
u(k) \\
u(k+1) \\
u(k+2) \\
\vdots \\
u(k+N_c-1)
\end{bmatrix}
\]
\vspace{-8pt}
\[
\Theta = \begin{bmatrix} 
B & 0 & 0 & \cdots & 0 \\
AB & B & 0 & \cdots & 0 \\
A^2B & AB & B & \cdots & 0 \\
\vdots & \vdots & \vdots & \ddots & \vdots \\
A^{N_c-1}B & A^{N_c-2}B & A^{N_c-3}B & \cdots & 0 \\
\vdots & \vdots & \vdots & \ddots & \vdots \\
A^{N_p-1}B & A^{N_p-2}B & A^{N_p-3}B & \cdots & A^{N_p-N_c+1}B
\end{bmatrix}
\]
To ensure that the vehicle follows the expected trajectory \(Y_t\), while maintaining driving stability, the objective function is defined as:
\begin{equation}
J = (Y - Y_t)^T Q_Q (Y - Y_t) + U^T R_R U,
\label{eq:5}
\end{equation}
where \( Q_Q \) and \( R_R \) are weight matrices.
By combining Equation (\ref{eq:4}) and Equation (\ref{eq:5}), we obtain:
\begin{equation}
J = 2\left( \frac{1}{2} U^T H U + g^T U \right) + \mathbb{C},
\label{eq:27}
\end{equation}
where \( \mathbb{C} \) is a constant, and \( H \) and \( g \) are expressed as follows:
\[
\begin{aligned}
H &= \Theta^T Q_Q \Theta + R_R, \\
g &= \Theta^T Q_Q (\Psi \mathbf{X}(k) - Y_t).
\end{aligned}
\]
To ensure the stability of the vehicle, the ranges of \( U \) and \( \Delta U \) must be limited, i.e., \( U_{\min} < U < U_{\max} \), and \( \Delta U_{\min} < \Delta U < \Delta U_{\max} \). The relationship between \(U\) and \( \Delta U\) is defined as follows:
\begin{equation}
\Delta U = E_{N_c}^{-1}(U - U_{k-1}),
\label{eq:6}
\end{equation}
And, \(E_{N_c}\), \(U_{k-1}\) are expressed as follows:
\[
\Delta U = 
\begin{bmatrix}
\Delta u(k) \\
\Delta u(k+1) \\
\vdots \\
\Delta u(k+N_c-1)
\end{bmatrix}
, \quad
U_{k-1} = 
\begin{bmatrix}
u(k-1) \\
u(k-1) \\
\vdots \\
u(k-1)
\end{bmatrix},
\]
\[
E_{N_c} = 
\begin{bmatrix}
E_3 & 0 & \cdots & 0 \\
E_3 & E_3 & \cdots & 0 \\
\vdots & \vdots & \ddots & 0 \\
E_3 & E_3 & \cdots & E_3
\end{bmatrix}.
\]
In summary, the path tracking problem has now been transformed into a quadratic programming problem:
\begin{equation}
\begin{split}
\min_U J &= \frac{1}{2} U^T H U + g^T U ,\\
\text{s.t.}
&\left\{
\begin{aligned}
&U_{\min} \leq U \leq U_{\max} \\
&\Delta U_{\min} \leq \Delta U \leq \Delta U_{\max}
\end{aligned}
\right.
\end{split}
\label{eq:7}
\end{equation}
By optimizing the objective function \( J \), the optimal value of \( U \) under the constraint conditions can be obtained. Consequently, the vehicle's optimal control vector \( u_{\text{best}} = [V_{x}, V_{y}, \omega]^T \) is determined. Next, the control vector will be used to calculate the rotational speed and steering angle of each wheel.
\begin{figure}
    \centering
    \includegraphics[width=1\linewidth]{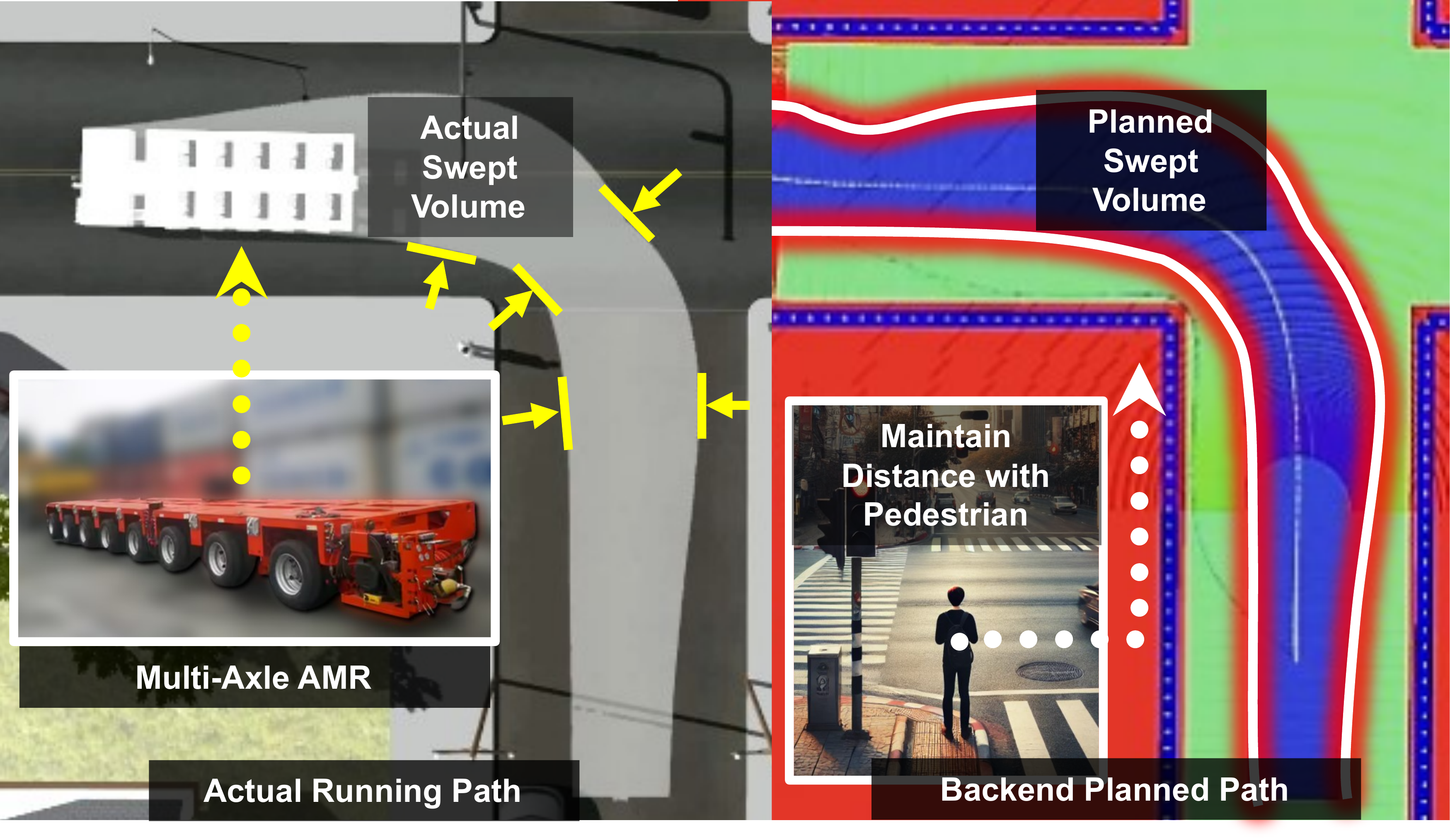}
    \vspace{-17pt}
    \caption{\footnotesize Experimental results show the proposed MPC accurately tracks the planned trajectory while minimizing robot travel in LiDAR blind spots. }
    \label{fig:simulation}  \vspace{-11pt}
\end{figure}
\begin{figure}
    \centering
    \includegraphics[width=1\linewidth]{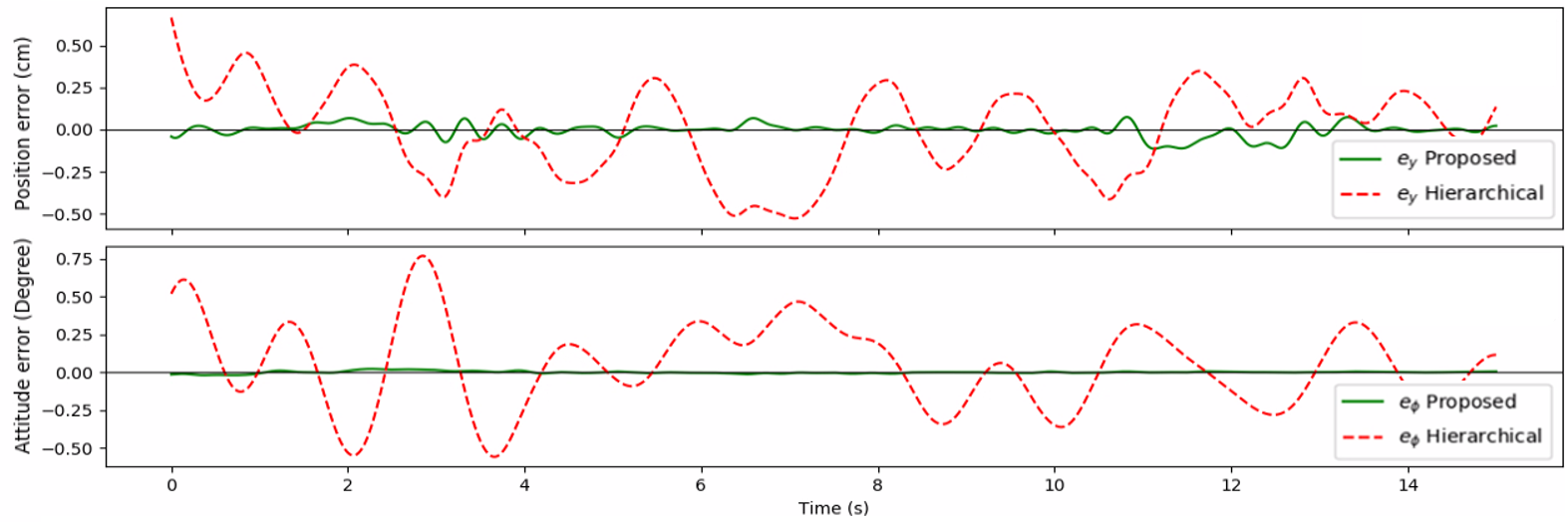}
    \vspace{-20pt}
    \caption{\footnotesize Trajectory tracking error comparison.}
    \vspace{-15pt}
    \label{fig:Trajectory tracking error comparison}
\end{figure}
\begin{figure*}
    \centering
    \includegraphics[width=0.95\linewidth]{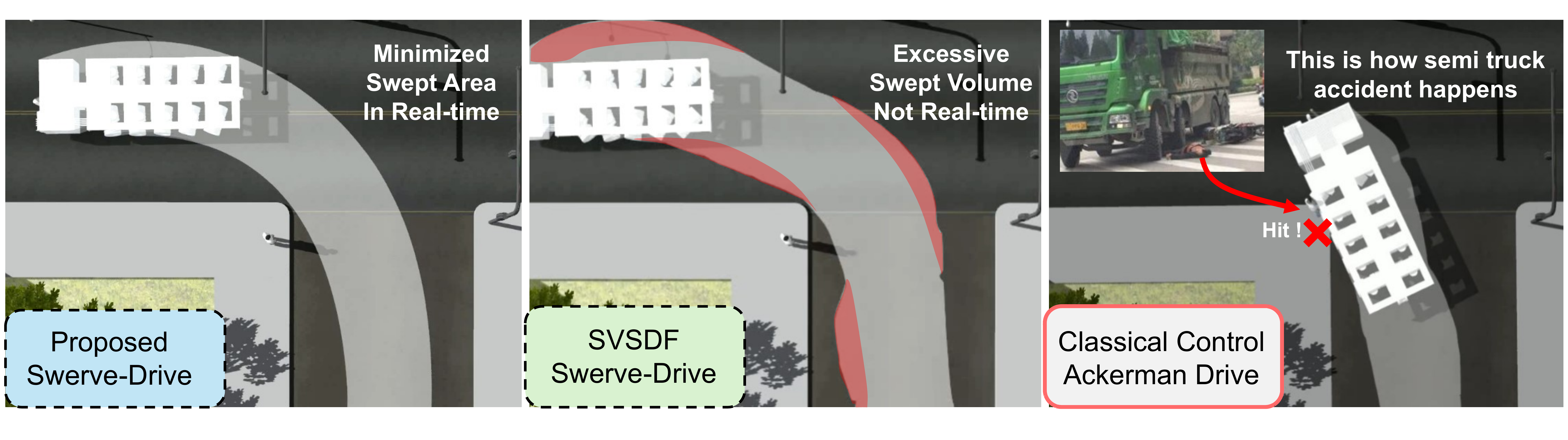}
    \vspace{-15pt}
    \caption{\footnotesize Experiment Result, Proposed solution offers minimal swept volume compared to another baseline model. }
    \label{fig:simulation1}
    \vspace{-13pt}
\end{figure*}

As shown in Fig. \ref{fig:modeling}, in the vehicle coordinate system, the rotational velocity  \(\omega\) of the vehicle can be decomposed to the X-axis and Y-axis are represented as \(V_{\tau ix}\) and \(V_{\tau iy}\), respectively. Therefore, the resultant velocity of the wheel \(\vec{V}_i = [V_{ix}, V_{iy}]^T \) can be expressed as:
\begin{equation}
\vec{V}_i = 
\begin{bmatrix}
V_{ix} \\
V_{iy}
\end{bmatrix}
=
\begin{bmatrix}
V_x + V_{\tau ix} \\
V_y + V_{\tau iy}
\end{bmatrix}
=
\begin{bmatrix}
1 & 0 & Y_{wi} \\
0 & 1 & X_{wi}
\end{bmatrix}
\begin{bmatrix}
V_x \\
V_y \\
\omega
\end{bmatrix}.
\label{eq:1}
\end{equation}
Therefore, the final steering angles \(\gamma_{i}\) and the linear speed \(V_{i}\) of each wheel group are expressed as follows:
\begin{equation}
\gamma_{i} = \arctan\left(\frac{V_{iy}}{V_{ix}}\right),
\label{eq:2}
\end{equation}
\begin{equation}
V_{i} = \sqrt{(V_{ix})^2 + (V_{iy})^2}.
\label{eq:speed}
\end{equation}
By combining Equations (\ref{eq:1}),  (\ref{eq:2}), and (\ref{eq:speed}), the vehicle's point mass control vector \( u_{\text{best}} \) can be transformed into the linear speed \(V{i}\) and steering angle \(\gamma_{i}\) for each wheel group, thereby controlling the vehicle to follow the target trajectory.

\section{Experiment}
\subsection{System Setup}
This work is designed for multi-axle swerve-drive AMRs in future logistics. However, due to hardware limitations, we rely on simulations to verify performance. As shown in Fig.~\ref{fig:simulation}, a street map is set up in Gazebo, where a pedestrian is standing at a crosswalk, preparing to cross the road. A 5-axle Swerve-Drive Vehicle, approximately 8.1m long and 2.7m wide, equipped with a LiDAR sensor in front, is making a left turn at the intersection. The objective is to plan a path that minimizes the $\mathbb{S}$ while keeping the Swept area away from obstacles to ensure pedestrian safety. The path is tracked using an MPC controller.
All the experiments are done on a Gen 13 i7 Notebook PC with Nvidia 4060 GPU.

\subsection{Evaluation Metric}
The evaluation metric for our work is listed as follows:
\begin{itemize}
    \item \textbf{Excess Swept Area $\mathbb{S}_{\text{excess}}$:} The additional volume covered by the vehicle beyond the minimal swept area, defined as the difference between the actual and minimal swept area, as illustrated in Fig.  \ref{fig:firstpage}.%:
    % \[
    % \mathbb{S}_{\text{excess}} = \mathbb{S}_{\text{actual}} - \mathbb{S}_{\text{minimal}}
    % \]
    % where $\mathbb{S}_{\text{minimal}}$ is the minimal area the vehicle would sweep if following its shortest path, plus the area of the vehicle itself. Minimizing $\mathbb{S}_{\text{excess}}$ reduces the risk of obstacles
    \item \textbf{Planning Time $t$:} Quick trajectory generation is vital for real-time applications.
    \item \textbf{Tracking Error $e_y$, $e_{\varphi}$:} Accurate tracking ensures safe and efficient path execution.
\end{itemize}

\subsection{Result and Discussion}
\begin{table}[t]
\centering
\renewcommand{\arraystretch}{1.3} % Increase row spacing
\caption{Comparison of Metrics for Different Methods}
\begin{tabular}{lccccc}
\hline
\hline
\textbf{Method} & \textbf{$\mathbb{S}_{\text{excess}}$ ($m^2$)} & \textbf{$t$ ($s$)} & \textbf{$e_y$ ($m$)} & \textbf{$e_{\varphi}$ (°)} \\ \hline 
 Classic\cite{zhao2019modelling} & 82.15                      & 1.50                                 & ±1.38                       & ±3.6\\ \hline
SVSDF \cite{wang2024implicit}        & 75.62                      & 2.5                                  & ±0.04                       & ±0.04 \\ \hline
Hierarchical \cite{xu2022hierarchical}       & 48.37                        & 1.38                                  & ±0.13                       & ±0.12 \\ \hline
Proposed & \textbf{23.14}                      & \textbf{1.17}                                 & ±\textbf{0.04}                       & ±\textbf{0.03 }\\ \hline
\end{tabular}
\label{tab:comparison}
\vspace{3pt}
\footnotesize{Classic~\cite{zhao2019modelling} and Hierarchical \cite{xu2022hierarchical} are designed solely for tracking the trajectory generated by our proposed multi-axle AMR planner. SVSDF~\cite{wang2024implicit} only has planning, and control is done using our proposed MPC .}
\vspace{-15pt}
\end{table}

\textbf{Table~\ref{tab:comparison}} and \textbf{Fig.~\ref{fig:Trajectory tracking error comparison}} present a comparison of various methods based on selected evaluation metrics. The Classic method refers to the trajectory tracking approach proposed by~\cite{zhao2019modelling}, where we set the model to allow only the front wheels to steer, simulating a traditional truck. The Hierarchical method is based on the trajectory tracking approach from~\cite{xu2022hierarchical}. The SVSDF method corresponds to the trajectory planning approach from~\cite{wang2024implicit}, and it is designed to use our MPC controller to track the generated trajectory. Both the Classic and Hierarchical methods are designed solely for tracking the trajectory generated by our proposed multi-axle AMR planner. For the Hierarchical and SVSDF methods, we use a Multi-Axle Swerve-Drive AMR model, where all-wheel groups are steerable.

The proposed method achieves the smallest $\mathbb{S}_{\text{excess}}$, measuring 23.14 m$^2$, which is the minimal excessive swept area by the vehicle.
This is crucial for avoiding obstacles and ensuring pedestrian safety. Although the Classic and Hierarchical methods track the trajectory generated by our proposed approach, their poor tracking performance results in a significantly larger $\mathbb{S}_{\text{excess}}$ making them much less effective compared to our method.
In terms of trajectory planning time, our method significantly outperforms others, with a reduced planning time of 1.17 seconds. The Classic and Hierarchical methods are designed to use the trajectory generated by our proposed approach, which results in similarly fast trajectory generation times. This improvement is critical for real-time applications where rapid responses to environmental changes are required.

For trajectory tracking performance, our approach maintains the lateral tracking error, \( e_y \), within ±0.04 m and the heading angle error, \( e_{\varphi} \), within ±0.03°. Compared to the Hierarchical method, our approach demonstrates substantial improvements in both precision and reliability, thereby enhancing the safety and maneuverability of the vehicle. Since the SVSDF method is designed to use our proposed tracking method, it achieves similar accuracy in tracking error performance.

%We also compare with different AMR models as shown in Fig. \ref{fig:simulation1}. The proposed methods show minimal swept areas without collisions for the newer type of AMR. While SVSDF reduces crash risk, its swept volume is suboptimal. The classical truck control method performs the worst, often causing collisions, as seen in many news reports.
We conducted a comparative analysis with different AMR models, as shown in Fig. \ref{fig:simulation1}. The proposed methods achieve minimal swept areas and accurate trajectory tracking without collisions, providing significant improvements for modern AMRs. While SVSDF reduces collision risk, its swept volume is still suboptimal. The classical truck control method performed the worst, with the largest swept area and frequent collisions, as commonly reported in industrial settings and news sources. This highlights the effectiveness of our approach in minimizing swept volume and enhancing safety.
%????? no Chinese. there is bug with chinese 
% OK! HA HA

\section{Conclusion}
This paper presented a unified approach combining swept volume-aware path planning with MPC to optimize trajectories and independently control each wheel of multi-axle AMRs for precise maneuvering in constrained environments. By calculating wheel group steering angles from velocity vectors, we simplify the vehicle model and enhance MPC control. Simulations show reduced swept volume and improved real-time trajectory tracking using CUDA, supporting reliable and efficient autonomous heavy-duty AMR applications. %We plan to open-source our work for the benefit of the community.

\newpage

\bibliographystyle{IEEEtran}
%\bibliography{referernces}
\bibliography{mybib}

\begin{thebibliography}{10}
\providecommand{\url}[1]{#1}
\csname url@samestyle\endcsname
\providecommand{\newblock}{\relax}
\providecommand{\bibinfo}[2]{#2}
\providecommand{\BIBentrySTDinterwordspacing}{\spaceskip=0pt\relax}
\providecommand{\BIBentryALTinterwordstretchfactor}{4}
\providecommand{\BIBentryALTinterwordspacing}{\spaceskip=\fontdimen2\font plus
\BIBentryALTinterwordstretchfactor\fontdimen3\font minus \fontdimen4\font\relax}
\providecommand{\BIBforeignlanguage}[2]{{%
\expandafter\ifx\csname l@#1\endcsname\relax
\typeout{** WARNING: IEEEtran.bst: No hyphenation pattern has been}%
\typeout{** loaded for the language `#1'. Using the pattern for}%
\typeout{** the default language instead.}%
\else
\language=\csname l@#1\endcsname
\fi
#2}}
\providecommand{\BIBdecl}{\relax}
\BIBdecl

\bibitem{zhao2019modelling}
H.~Zhao, Z.~Liu, Z.~Li, S.~Zhou, W.~Chen, C.~Suo, and Y.-H. Liu, ``Modelling and dynamic tracking control of industrial vehicles with tractor-trailer structure,'' in \emph{2019 IEEE/RSJ International Conference on Intelligent Robots and Systems (IROS)}.\hskip 1em plus 0.5em minus 0.4em\relax IEEE, 2019.

\bibitem{islam2020planning}
F.~Islam, A.~Vemula, S.-K. Kim, A.~Dornbush, O.~Salzman, and M.~Likhachev, ``Planning, learning and reasoning framework for robot truck unloading,'' in \emph{2020 IEEE International Conference on Robotics and Automation (ICRA)}.\hskip 1em plus 0.5em minus 0.4em\relax IEEE, 2020, pp. 5011--5017.

\bibitem{yuan2021survey}
S.~Yuan, H.~Wang, and L.~Xie, ``Survey on localization systems and algorithms for unmanned systems,'' \emph{Unmanned Systems}, vol.~9, no.~02, pp. 129--163, 2021.

\bibitem{zhang2023continuous}
T.~Zhang, J.~Wang, C.~Xu, A.~Gao, and F.~Gao, ``Continuous implicit sdf based any-shape robot trajectory optimization,'' in \emph{2023 IEEE/RSJ International Conference on Intelligent Robots and Systems (IROS)}.\hskip 1em plus 0.5em minus 0.4em\relax IEEE, 2023, pp. 282--289.

\bibitem{wang2024implicit}
J.~Wang, T.~Zhang, Q.~Zhang, C.~Zeng, J.~Yu, C.~Xu, L.~Xu, and F.~Gao, ``Implicit swept volume sdf: Enabling continuous collision-free trajectory generation for arbitrary shapes,'' \emph{ACM Transactions on Graphics (TOG)}, vol.~43, no.~4, pp. 1--14, 2024.

\bibitem{liu2020experimental}
X.~Liu, A.~K. Madhusudhanan, and D.~Cebon, ``Experimental evaluation of minimum swept-path control for autonomous reversing of articulated vehicles,'' in \emph{2020 IEEE Intelligent Vehicles Symposium (IV)}.\hskip 1em plus 0.5em minus 0.4em\relax IEEE, 2020, pp. 584--589.

\bibitem{yang2024e2e}
Y.~Yang, D.~Chen, T.~Qin, X.~Mu, C.~Xu, and M.~Yang, ``E2e parking: Autonomous parking by the end-to-end neural network on the carla simulator,'' in \emph{2024 IEEE Intelligent Vehicles Symposium (IV)}.\hskip 1em plus 0.5em minus 0.4em\relax IEEE, 2024, pp. 2375--2382.

\bibitem{bai2021multi}
R.~Bai, R.~Zheng, M.~Liu, and S.~Zhang, ``Multi-robot task planning under individual and collaborative temporal logic specifications,'' in \emph{2021 IEEE/RSJ International Conference on Intelligent Robots and Systems (IROS)}.\hskip 1em plus 0.5em minus 0.4em\relax IEEE, 2021, pp. 6382--6389.

\bibitem{bian2023risk}
J.~Bian, J.~Zhang, K.~Guo, W.~Li, X.~Yu, and L.~Guo, ``Risk-aware path planning using cvar for quadrotors,'' in \emph{2023 6th International Symposium on Autonomous Systems (ISAS)}.\hskip 1em plus 0.5em minus 0.4em\relax IEEE, 2023, pp. 1--6.

\bibitem{jin2024gs}
R.~Jin, Y.~Gao, Y.~Wang, Y.~Wu, H.~Lu, C.~Xu, and F.~Gao, ``Gs-planner: A gaussian-splatting-based planning framework for active high-fidelity reconstruction,'' in \emph{2024 IEEE/RSJ International Conference on Intelligent Robots and Systems (IROS)}.\hskip 1em plus 0.5em minus 0.4em\relax IEEE, 2024, pp. 11\,202--11\,209.

\bibitem{liu2023safe}
S.~Liu, K.~Guo, X.~Yu, L.~Ma, L.~Xie, and L.~Guo, ``Safe maneuvering planning for flights in complex environments,'' \emph{IEEE Transactions on Industrial Electronics}, vol.~71, no.~5, pp. 4944--4953, 2023.

\bibitem{cao2023path}
M.~Cao, K.~Cao, S.~Yuan, K.~Liu, Y.~L. Wong, and L.~Xie, ``Path planning for multiple tethered robots using topological braids,'' in \emph{Robotics: Science and Systems}, 2023.

\bibitem{cao2023neptune}
M.~Cao, K.~Cao, S.~Yuan, T.-M. Nguyen, and L.~Xie, ``Neptune: non-entangling trajectory planning for multiple tethered unmanned vehicles,'' \emph{IEEE Transactions on Robotics}, vol.~39, no.~4, pp. 2786--2804, 2023.

\bibitem{li2024jacquard}
Q.~Li and S.~Yuan, ``Jacquard v2: Refining datasets using the human in the loop data correction method,'' in \emph{2024 IEEE International Conference on Robotics and Automation (ICRA)}, 2024, pp. 7932--7938.

\bibitem{zhao2024design}
C.~Zhao, E.~Wang, K.~Guo, and X.~Yu, ``Design and anti-disturbance control for an h-configuration tiltable quadrotor: Enabling narrow space crossing,'' \emph{IEEE Transactions on Aerospace and Electronic Systems}, 2024.

\bibitem{yang2024trace}
Z.~Yang, J.~Jia, Y.~Liu, K.~Guo, X.~Yu, and L.~Guo, ``Trace: Trajectory refinement with control error enables safe and accurate maneuvers,'' in \emph{2024 IEEE 18th International Conference on Control \& Automation (ICCA)}.\hskip 1em plus 0.5em minus 0.4em\relax IEEE, 2024, pp. 154--161.

\bibitem{fan2024flying}
D.~Fan, Q.~Liu, C.~Zhao, K.~Guo, Z.~Yang, X.~Yu, and L.~Guo, ``Flying in narrow spaces: Prioritizing safety with disturbance-aware control,'' \emph{IEEE Robotics and Automation Letters}, 2024.

\bibitem{yu2024aggressive}
H.~Yu, C.~Hu, J.~Wang, G.~Lu, J.~Tu, Z.~Zheng, J.~Li, and F.~Gao, ``Aggressive collision-inclusive motion planning,'' \emph{IEEE/ASME Transactions on Mechatronics}, 2024.

\bibitem{Bai2025Realm}
R.~Bai, S.~Yuan, K.~Li, H.~Guo, W.-Y. Yau, and L.~Xie, ``Realm: Real-time line-of-sight maintenance in multi-robot navigation with unknown obstacles,'' in \emph{IEEE International Conference on Robotics and Automation (ICRA)}, 2025.

\bibitem{cao2019bearing}
K.~Cao, D.~Li, and L.~Xie, ``Bearing-ratio-of-distance rigidity theory with application to directly similar formation control,'' \emph{Automatica}, vol. 109, p. 108540, 2019.

\bibitem{liu2023non}
F.~Liu, S.~Yuan, W.~Meng, R.~Su, and L.~Xie, ``Non-cooperative stochastic target encirclement by anti-synchronization control via range-only measurement,'' in \emph{2023 IEEE International Conference on Robotics and Automation (ICRA)}.\hskip 1em plus 0.5em minus 0.4em\relax IEEE, 2023, pp. 5480--5485.

\bibitem{hu2023stackelberg}
Z.~Hu, X.~Li, and M.~Meng, ``A stackelberg game approach for three-player autonomous racing,'' in \emph{2023 42nd Chinese Control Conference (CCC)}.\hskip 1em plus 0.5em minus 0.4em\relax IEEE, 2023, pp. 4886--4891.

\bibitem{cao2020online}
M.~Cao, Y.~Lyu, S.~Yuan, and L.~Xie, ``Online trajectory correction and tracking for facade inspection using autonomous uav,'' in \emph{2020 IEEE 16th International Conference on Control \& Automation (ICCA)}.\hskip 1em plus 0.5em minus 0.4em\relax IEEE, 2020, pp. 1149--1154.

\bibitem{liu2023multiple}
F.~Liu, S.~Yuan, W.~Meng, R.~Su, and L.~Xie, ``Multiple noncooperative targets encirclement by relative distance-based positioning and neural antisynchronization control,'' \emph{IEEE Transactions on Industrial Electronics}, vol.~71, no.~2, pp. 1675--1685, 2023.

\bibitem{xu2024cost}
X.~Xu, M.~Cao, S.~Yuan, T.~H. Nguyen, T.-M. Nguyen, and L.~Xie, ``A cost-effective cooperative exploration and inspection strategy for heterogeneous aerial system,'' in \emph{Proceedings of the 2024 IEEE International Conference on Control and Automation (ICCA)}.\hskip 1em plus 0.5em minus 0.4em\relax IEEE, 2024.

\bibitem{er2013development}
M.~J. Er, S.~Yuan, and N.~Wang, ``Development control and navigation of octocopter,'' in \emph{2013 10th IEEE International Conference on Control and Automation (ICCA)}.\hskip 1em plus 0.5em minus 0.4em\relax IEEE, 2013, pp. 1639--1643.

\bibitem{jia2023evolver}
J.~Jia, W.~Zhang, K.~Guo, J.~Wang, X.~Yu, Y.~Shi, and L.~Guo, ``Evolver: Online learning and prediction of disturbances for robot control,'' \emph{IEEE Transactions on Robotics}, 2023.

\bibitem{liu2024distance}
F.~Liu, S.~Yuan, K.~Cao, W.~Meng, and L.~Xie, ``Distance-based multiple noncooperative ground target encirclement for complex environments,'' \emph{IEEE Transactions on Control Systems Technology}, vol.~33, no.~1, pp. 261--273, 2025.

\bibitem{ji2024integration}
M.~Ji, K.~Pan, X.~Zhang, Q.~Pan, X.~Dai, and Y.~Lyu, ``Integration of sense and control for uncertain systems based on delayed feedback active inference,'' \emph{Entropy}, vol.~26, no.~11, p. 990, 2024.

\bibitem{lee2022cyclops}
H.~Lee, J.~Park, C.~Koo, J.-C. Kim, and Y.~Eun, ``Cyclops: Open platform for scale truck platooning,'' in \emph{2022 International Conference on Robotics and Automation (ICRA)}.\hskip 1em plus 0.5em minus 0.4em\relax IEEE, 2022, pp. 8971--8977.

\bibitem{nguyen2024eigen}
T.-M. Nguyen, X.~Xu, T.~Jin, Y.~Yang, J.~Li, S.~Yuan, and L.~Xie, ``Eigen is all you need: Efficient lidar-inertial continuous-time odometry with internal association,'' \emph{IEEE Robotics and Automation Letters}, 2024.

\bibitem{baxter2020deep}
J.~Baxter, M.~R. Yousefi, S.~Sugaya, M.~Morales, and L.~Tapia, ``Deep prediction of swept volume geometries: Robots and resolutions,'' in \emph{2020 IEEE/RSJ International Conference on Intelligent Robots and Systems (IROS)}.\hskip 1em plus 0.5em minus 0.4em\relax IEEE, 2020, pp. 6665--6672.

\bibitem{chiang2021fast}
H.-T.~L. Chiang, J.~E. Baxter, S.~Sugaya, M.~R. Yousefi, A.~Faust, and L.~Tapia, ``Fast deep swept volume estimator,'' \emph{The International Journal of Robotics Research}, vol.~40, no. 10-11, pp. 1068--1086, 2021.

\bibitem{zips2015optimisation}
P.~Zips, M.~B{\"o}ck, and A.~Kugi, ``An optimisation-based path planner for truck-trailer systems with driving direction changes,'' in \emph{2015 IEEE International Conference on Robotics and Automation (ICRA)}.\hskip 1em plus 0.5em minus 0.4em\relax IEEE, 2015, pp. 630--636.

\bibitem{cao2019preview}
K.~Cao, X.~Li, and L.~Xie, ``Preview-based discrete-time dynamic formation control over directed networks via matrix-valued laplacian,'' \emph{IEEE Transactions on Cybernetics}, vol.~50, no.~3, 2019.

\bibitem{ilic2018vehicle}
V.~Ili{\'c}, D.~Gavran, S.~Fric, F.~Trp{\v{c}}evski, and S.~Vranjevac, ``Vehicle swept path analysis based on gps data,'' \emph{Canadian Journal of Civil Engineering}, vol.~45, no.~10, pp. 827--839, 2018.

\bibitem{chen2024salient}
S.~Chen, K.~Liu, C.~Wang, S.~Yuan, J.~Yang, and L.~Xie, ``Salient sparse visual odometry with pose-only supervision,'' \emph{IEEE Robotics and Automation Letters}, vol.~9, no.~5, pp. 4774--4781, 2024.

\bibitem{esfahani2019deepdsair}
M.~A. Esfahani, K.~Wu, S.~Yuan, and H.~Wang, ``Deepdsair: Deep 6-dof camera relocalization using deblurred semantic-aware image representation for large-scale outdoor environments,'' \emph{Image and Vision Computing}, vol.~89, pp. 120--130, 2019.

\bibitem{yang2024fast}
Z.~Yang, K.~Xu, S.~Yuan, and L.~Xie, ``A fast and light-weight noniterative visual odometry with rgb-d cameras,'' \emph{Unmanned Systems}, vol.~0, no.~0, pp. 1--13, 2024.

\bibitem{10612831}
J.~Xu, W.~Yu, S.~Huang, S.~Yuan, L.~Zhao, R.~Li, and L.~Xie, ``M-divo: Multiple tof rgb-d cameras-enhanced depth–inertial–visual odometry,'' \emph{IEEE Internet of Things Journal}, vol.~11, no.~23, pp. 37\,562--37\,570, 2024.

\bibitem{ji2022robust}
T.~Ji, S.~Yuan, and L.~Xie, ``Robust rgb-d slam in dynamic environments for autonomous vehicles,'' in \emph{2022 17th International Conference on Control, Automation, Robotics and Vision (ICARCV)}.\hskip 1em plus 0.5em minus 0.4em\relax IEEE, 2022, pp. 665--671.

\bibitem{Li2024graph}
J.~Li, T.-M. Nguyen, M.~Cao, S.~Yuan, T.-Y. Hung, and L.~Xie, ``Graph optimality-aware stochastic lidar bundle adjustment with progressive spatial smoothing,'' in \emph{arXiv preprint arXiv:2410.14565}, 2024.

\bibitem{esfahani2018new}
M.~A. Esfahani, K.~Wu, S.~Yuan, and H.~Wang, ``A new approach to train convolutional neural networks for real-time 6-dof camera relocalization,'' in \emph{2018 IEEE 14th international conference on control and automation (ICCA)}.\hskip 1em plus 0.5em minus 0.4em\relax IEEE, 2018, pp. 81--85.

\bibitem{10801455}
W.~Yu, J.~Xu, C.~Zhao, L.~Zhao, T.-M. Nguyen, S.~Yuan, M.~Bai, and L.~Xie, ``I2ekf-lo: A dual-iteration extended kalman filter based lidar odometry,'' in \emph{2024 IEEE/RSJ International Conference on Intelligent Robots and Systems (IROS)}, 2024, pp. 10\,453--10\,460.

\bibitem{esfahani2020local}
M.~A. Esfahani, K.~Wu, S.~Yuan, and H.~Wang, ``From local understanding to global regression in monocular visual odometry,'' \emph{International Journal of Pattern Recognition and Artificial Intelligence}, vol.~34, no.~01, p. 2055002, 2020.

\bibitem{10802614}
J.~Li, T.-M. Nguyen, S.~Yuan, and L.~Xie, ``Pss-ba: Lidar bundle adjustment with progressive spatial smoothing,'' in \emph{2024 IEEE/RSJ International Conference on Intelligent Robots and Systems (IROS)}, 2024, pp. 1124--1129.

\bibitem{jin2024robust}
T.~Jin, X.~Xu, Y.~Yang, S.~Yuan, T.-M. Nguyen, J.~Li, and L.~Xie, ``Robust loop closure by textual cues in challenging environments,'' \emph{IEEE Robotics and Automation Letters}, vol.~10, no.~1, pp. 812--819, 2025.

\bibitem{10802691}
R.~Bai, S.~Yuan, H.~Guo, P.~Yin, W.-Y. Yau, and L.~Xie, ``Multi-robot active graph exploration with reduced pose-slam uncertainty via submodular optimization,'' in \emph{2024 IEEE/RSJ International Conference on Intelligent Robots and Systems (IROS)}, 2024, pp. 10\,229--10\,236.

\bibitem{li2024hcto}
J.~Li, S.~Yuan, M.~Cao, T.-M. Nguyen, K.~Cao, and L.~Xie, ``Hcto: Optimality-aware lidar inertial odometry with hybrid continuous time optimization for compact wearable mapping system,'' \emph{ISPRS Journal of Photogrammetry and Remote Sensing}, vol. 211, pp. 228--243, 2024.

\bibitem{Nguyen2025ULOC}
T.-M. Nguyen, Y.~Yang, T.-D. Nguyen, S.~Yuan, and L.~Xie, ``Uloc: Learning to localize in complex large-scale environments with ultra-wideband ranges,'' in \emph{IEEE International Conference on Robotics and Automation (ICRA)}, 2025.

\bibitem{li2024ua}
J.~Li, X.~Xu, J.~Liu, K.~Cao, S.~Yuan, and L.~Xie, ``Ua-mpc: Uncertainty-aware model predictive control for motorized lidar odometry,'' in \emph{arXiv preprint arXiv:2412.13873}, 2024.

\bibitem{yin2024outram}
P.~Yin, H.~Cao, T.-M. Nguyen, S.~Yuan, S.~Zhang, K.~Liu, and L.~Xie, ``Outram: One-shot global localization via triangulated scene graph and global outlier pruning,'' in \emph{2024 IEEE International Conference on Robotics and Automation (ICRA)}.\hskip 1em plus 0.5em minus 0.4em\relax IEEE, 2024, pp. 13\,717--13\,723.

\bibitem{ji2024sgba}
X.~Ji, S.~Yuan, J.~Li, P.~Yin, H.~Cao, and L.~Xie, ``Sgba: Semantic gaussian mixture model-based lidar bundle adjustment,'' \emph{IEEE Robotics and Automation Letters}, vol.~9, no.~12, pp. 10\,922--10\,929, 2024.

\bibitem{esfahani2021learning}
M.~A. Esfahani, H.~Wang, B.~Bashari, K.~Wu, and S.~Yuan, ``Learning to extract robust handcrafted features with a single observation via evolutionary neurogenesis,'' \emph{Applied Soft Computing}, vol. 106, p. 107424, 2021.

\bibitem{esfahani2019towards}
M.~A. Esfahani, K.~Wu, S.~Yuan, and H.~Wang, ``Towards utilizing deep uncertainty in traditional slam,'' in \emph{2019 IEEE 15th International Conference on Control and Automation (ICCA)}, 2019, pp. 344--349.

\bibitem{lyu2023spins}
Y.~Lyu, T.-M. Nguyen, L.~Liu, M.~Cao, S.~Yuan, T.~H. Nguyen, and L.~Xie, ``Spins: A structure priors aided inertial navigation system,'' \emph{Journal of Field Robotics}, vol.~40, no.~4, pp. 879--900, 2023.

\bibitem{Täubigsweptvolume}
H.~Täubig, B.~Bäuml, and U.~Frese, ``Real-time swept volume and distance computation for self collision detection,'' in \emph{2011 IEEE/RSJ International Conference on Intelligent Robots and Systems}, 2011, pp. 1585--1592.

\bibitem{yuan2014Autonomous}
S.~Yuan and H.~Wang, ``Autonomous object level segmentation,'' in \emph{Proceedings of International Conference on Control, Automation, Robotics and Vision (ICARCV 2014)}, 2014, pp. 33--37.

\bibitem{wang2015automatic}
H.~Wang, W.~Mou, X.~Mou, S.~Yuan, S.~Ulun, S.~Yang, and B.-S. Shin, ``An automatic self-calibration approach for wide baseline stereo cameras using sea surface images,'' \emph{Unmanned Systems}, vol.~3, no.~04, pp. 277--290, 2015.

\bibitem{wang2017heterogeneous}
H.~Wang, S.~Yuan, and K.~Wu, ``Heterogeneous stereo: A human vision inspired method for general robotics sensing,'' in \emph{TENCON 2017-2017 IEEE Region 10 Conference}.\hskip 1em plus 0.5em minus 0.4em\relax IEEE, 2017, pp. 793--798.

\bibitem{wu2019depth}
K.~Wu, M.~A. Esfahani, S.~Yuan, and H.~Wang, ``Depth-based obstacle avoidance through deep reinforcement learning,'' in \emph{Proceedings of the 5th International Conference on Mechatronics and Robotics Engineering}, 2019, pp. 102--106.

\bibitem{WangSDF}
H.~Wang, Y.~Lin, W.~Zhang, W.~Ye, M.~Zhang, and X.~Dong, ``Safe autonomous exploration and adaptive path planning strategy using signed distance field,'' \emph{IEEE Access}, vol.~11, 2023.

\bibitem{lyu2021vision}
Y.~Lyu, M.~Cao, S.~Yuan, and L.~Xie, ``Vision based autonomous uav plane estimation and following for building inspection,'' in \emph{arXiv preprint arXiv:2102.01423}, 2021.

\bibitem{qi2024air}
Z.~Qi, S.~Yuan, F.~Liu, H.~Cao, T.~Deng, J.~Yang, and L.~Xie, ``Air-embodied: An efficient active 3dgs-based interaction and reconstruction framework with embodied large language model,'' in \emph{arXiv preprint arXiv:2409.16019}, 2024.

\bibitem{xu2022hierarchical}
F.-x. Xu, C.~Zhou, and X.-h. Liu, ``Hierarchical control strategies for multi-mode steering system of emergency rescue vehicle,'' \emph{Mechatronics}, vol.~85, p. 102834, 2022.

\bibitem{wang2023multi}
W.~Wang, J.~Li, X.~Li, Z.~Li, and N.~Guo, ``Multi-objective collaborative control method for multi-axle distributed vehicle assisted driving,'' \emph{Applied Sciences}, vol.~13, no.~13, p. 7769, 2023.

\bibitem{gao2014turning}
F.~Gao and X.-y. Li, ``Turning characteristic study of multi-axle compound steering vehicle,'' in \emph{2014 IEEE Conference and Expo Transportation Electrification Asia-Pacific (ITEC Asia-Pacific)}.\hskip 1em plus 0.5em minus 0.4em\relax IEEE, 2014, pp. 1--5.

\bibitem{zhang2022dual}
Z.~Zhang, X.-j. Ma, C.-g. Liu, and S.-g. Wei, ``Dual-steering mode based on direct yaw moment control for multi-wheel hub motor driven vehicles: Theoretical design and experimental assessment,'' \emph{Defence Technology}, vol.~18, no.~1, pp. 49--61, 2022.

\bibitem{hu2016control}
J.-b. HU, M.-m. FU, X.-y. LI, and J.~NI, ``Control strategy of the multi-axle distributed mechanic and electric drive vehicle's turning radius,'' \emph{Transactions of Beijing institute of Technology}, vol.~36, no.~11, pp. 1131--1135, 2016.

\bibitem{wu2017optimizing}
Z.~Wu and C.~Chen, ``On optimizing steering performance of multi-axle vehicle based on driving force control,'' in \emph{MATEC Web of Conferences}, vol. 124.\hskip 1em plus 0.5em minus 0.4em\relax EDP Sciences, 2017, p. 07005.

\bibitem{chaudhuri2009kinematic}
S.~Chaudhuri, V.~Saini, and M.~Singh, ``Kinematic analysis of multi-axle steering system for articulated vehicle,'' SAE Technical Paper, Tech. Rep., 2009.

\bibitem{wu2021learn}
K.~Wu, H.~Wang, M.~A. Esfahani, and S.~Yuan, ``Learn to navigate autonomously through deep reinforcement learning,'' \emph{IEEE Transactions on Industrial Electronics}, vol.~69, no.~5, pp. 5342--5352, 2021.

\bibitem{ye2016steering}
Y.~Ye, L.~He, and Q.~Zhang, ``Steering control strategies for a four-wheel-independent-steering bin managing robot,'' \emph{IFAC-PapersOnLine}, vol.~49, no.~16, pp. 39--44, 2016.

\bibitem{zhang2016study}
P.~Zhang, L.~Gao, and Y.~Zhu, ``Study on control schemes of flexible steering system of a multi-axle all-wheel-steering robot,'' \emph{Advances in Mechanical Engineering}, vol.~8, no.~6, p. 1687814016651556, 2016.

\bibitem{zhang2015steering}
P.~X. Zhang, L.~Gao, and Y.~Q. Zhu, ``The steering performance analysis of multi-axle vehicle based on sideslip angle control strategy,'' \emph{Applied Mechanics and Materials}, vol. 701, pp. 799--802, 2015.

\bibitem{zhang2015steering1}
P.~Zhang, Y.~Zhu, R.~Huang, and X.~Zhang, ``Steering mode research of five-axle all-wheel steering wheeled robots,'' \emph{Mechanical Design and Manufacturing}, no.~3, pp. 39--42, 2015.

\bibitem{wang2022geometrically}
Z.~Wang, X.~Zhou, C.~Xu, and F.~Gao, ``Geometrically constrained trajectory optimization for multicopters,'' \emph{IEEE Transactions on Robotics}, vol.~38, no.~5, pp. 3259--3278, 2022.

\bibitem{coppola2020lbfgs}
A.~Coppola and B.~M. Stewart, ``lbfgs: Efficient l-bfgs and owl-qn optimization in r,'' 2020.

\bibitem{karmitsa2020limited}
N.~Karmitsa, ``Limited memory bundle method and its variations for large-scale nonsmooth optimization,'' \emph{Numerical Nonsmooth Optimization: State of the Art Algorithms}, pp. 167--199, 2020.

\bibitem{armijo1966minimization}
L.~Armijo, ``Minimization of functions having lipschitz continuous first partial derivatives,'' \emph{Pacific Journal of mathematics}, vol.~16, no.~1, pp. 1--3, 1966.

\end{thebibliography}

% % %\vspace{12pt}
% % %\color{red}
% % %IEEE conference templates contain guidance text for composing and formatting conference papers. Please ensure that all template text is removed from your conference paper prior to submission to the conference. Failure to remove the template text from your paper may result in your paper not being published.

\end{document}